\documentclass[tinyml]{acmart}

\AtBeginDocument{%
  \providecommand\BibTeX{{%
    \normalfont B\kern-0.5em{\scshape i\kern-0.25em b}\kern-0.8em\TeX}}}

\setcopyright{rightsretained}
\copyrightyear{2025}
\acmYear{2025}

\usepackage{multirow}
\usepackage{enumitem}
\usepackage{url}
\usepackage{soul}

\begin{document}

\def\projname{SteROI-D}
\title{\projname: System Design and Mapping for Stereo Depth Inference on Regions of Interest}

\author{Jack Erhardt}
\email{erharj@umich.edu}
\affiliation{%
    \institution{University of Michigan}
    \city{Ann Arbor}
    \state{Michigan}
    \country{USA}
}

\author{Ziang Li}
\email{ziangli@umich.edu}
\affiliation{%
    \institution{University of Michigan}
    \city{Ann Arbor}
    \state{Michigan}
    \country{USA}
}

\author{Reid Pinkham}
\email{pinkhamr@meta.com}
\affiliation{%
    \institution{Reality Labs - Research}
    \city{Redmond}
    \state{Washington}
    \country{USA}
}

\author{Andrew Berkovich}
\email{andrew.berkovich@meta.com}
\affiliation{%
    \institution{Reality Labs - Research}
    \city{Redmond}
    \state{Washington}
    \country{USA}
}

\author{Zhengya Zhang}
\email{zhengya@umich.edu}
\affiliation{%
    \institution{University of Michigan}
    \city{Ann Arbor}
    \state{Michigan}
    \country{USA}
}

\renewcommand{\shortauthors}{Erhardt et al.}

\newcommand{\je}[1]{\textcolor{black}{#1}}
\newcommand{\zz}[1]{\textcolor{black}{#1}}
\newcommand{\ca}[1]{\textcolor{black}{#1}}

\begin{abstract}
    Machine learning algorithms have enabled high quality stereo depth estimation to run on Augmented and Virtual Reality (AR/VR) devices.
    However, high energy consumption across the full image processing stack prevents stereo depth algorithms from running effectively on battery-limited devices.
    This paper introduces \textit{\projname{}}, a full stereo depth system paired with a mapping methodology.
    \projname{} exploits Region-of-Interest (ROI) and temporal sparsity at the system level to save energy. \projname{}'s flexible and heterogeneous compute fabric supports diverse ROIs. Importantly, we introduce a systematic mapping methodology to effectively handle dynamic ROIs, thereby maximizing energy savings.
    Using these techniques, our 28nm prototype \projname{} design achieves up to 4.35$\times$ reduction in total system energy compared to a baseline ASIC.
\end{abstract}


\maketitle

\section{Introduction}\label{sec:intro}

Augmented and Virtual Reality (AR/VR) has made significant advancements in recent years in terms of quality and affordability \cite{meta_motiv1, meta_motiv2}, through the use of machine learning algorithms.
One crucial algorithm is depth estimation from stereo sensors, which plays a vital role in spatial computing, hand tracking \cite{depth_gesture_recog}, and 
passthrough rendering.
Conventional DNN based stereo depth algorithms use expensive hierarchical processing \cite{stereonet, mobilestereonet, hitnet}, 
\je{which are challenging to accelerate on power constrained platforms}.
Increased resolutions and frame rates in newer systems further increase these costs \cite{near_sensor_distributed}.
Consequently, accelerating these networks on AR/VR devices while meeting real-time latency requirements and operating within the energy budget of limited battery devices presents a challenge.

In this work, we propose \textit{\projname{}}, an AR/VR stereo depth system comprising a flexible architecture for processing dynamic Regions of Interest (ROIs), and a comprehensive mapping methodology to optimize ROI processing for energy efficiency while maintaining real-time performance. Our contributions are as follows:

\begin{itemize}[leftmargin=*]
    \item \je{The \textit{SteROI-D Algorithm}, which leverages Region-of-Interest (ROI) Sparsity to reduce per-frame depth extraction cost, and interleaved object detection and tracking to reduce ROI detection cost;}
    \item \je{Special Compute Units (SCUs) and NoC Multipackets, to address compute and communication challenges in accelerating stereo depth networks;}
    \item \je{\textit{Binned Mapping}, a method for split online-offline algorithm mapping to enable efficient processing for a continuous range of ROI sizes; and}
    \item \je{A design space exploration framework for jointly optimizing an accelerator's SRAM allocation with it's Binned Mapping.}
\end{itemize}

To our knowledge, this is the first study to achieve ROI-based stereo depth processing.
This is also the first work to address variable ROI processing through a mapping-system co-design approach via an efficient design space exploration.
While prior work has exploited ROIs for eye tracking on AR devices, it was limited to a static architecture \cite{eyecod}.
Furthermore, although prior work has also proposed lightweight stereo depth systems for AR devices, they have not leveraged ROI sparsity \cite{tiefenrausch}.
\section{Background}\label{sec:background}

\subsection{\je{Low Power Algorithms}}
Stereo depth processing consumes significant energy on AR/VR platforms.
For instance, on a Jetson Orin Nano, we measure stereo depth on a 90k pixel crop at 30~FPS consumes 5.6~W of power, or 400~mJ per inference.
As AR/VR devices employ higher resolution sensors to achieve more immersive experiences, it is anticipated that the computational intensity will escalate even more.

\subsection{Stereo Depth Processing}
Stereo depth has been the focus of many algorithmic works; as of writing, deep learning based approaches achieve the best inference quality. 
StereoNet \cite{stereonet} is an early attempt at an algorithm that can be accelerated on edge hardware in real time, and shares many foundational traits with subsequent networks. 
It uses twin Siamese feature extraction layers to initialize multi-resolution disparity estimates, which are hierarchically refined to produce a final disparity estimate. 
HITNet \cite{hitnet} iterates on this algorithm structure by introducing tile based iterative refinement, and local slant predictions alongside disparity estimation, to improve inference quality. 
This network also performs disparity processing without explicitly evaluating a 3D cost volume. 
More recently, the monocular depth network Tiefenrausch and it's stereo depth cousin Argos \cite{tiefenrausch} have been proposed using building blocks inspired by MobileNetv2 \cite{mobilenetv2}, and trained on 8-bit quantized weights for highly efficient inference.

\subsection{\je{AR Systems and Compute}}
\je{AR platforms incorporate many system and architectural techniques to achieve low power, low latency, and tight form factors.}
\je{Typical AR Systems on Chip (SoCs), such as the Qualcomm Snapdragon \cite{qualcomm_soc}, feature heterogeneously integrated accelerator, CPU, GPU, and memory units for diverse tasks.}
\je{Near-sensor computing \cite{eyecod, ansa, gomez_distributed_2022, sony_coprocessor, siracusa} is also commonly proposed as an energy-efficient computing technique for AR.}
\je{This is because near-sensor processing can be used to reduce raw sensor data sizes, saving expensive communication energy over MIPI or other protocols.}

\je{Several accelerator designs have been proposed for AR SoCs.}
\je{These accelerators must enable efficient inference at low batch sizes, while providing real-time latency, and small core areas to satisfy the stringent space limitations of AR platforms.}
\ca{For this work, we use a baseline real-time throughput and latency requirement of 30FPS, which is common for off-the-shelf image sensors and applications.}
\je{In particular, we consider the architecture proposed in ANSA \cite{ansa}, which enables efficient real-time processing with Vector-Matrix Multiplier (VMM) based compute units.}
\je{This architecture is also organized hierarchically and with parameterized scale, which enables design space exploration.}

\je{While this accelerator design works well for classification-based CNNs, stereo depth networks present new compute morphologies which must be addressed.}
\je{Modern stereo-depth models increasingly utilize non-parameterized special layer types, as illustrated in Figure ~\ref{fig:ops_compare}.}
\je{These operations cannot leverage VMM parallelism, and thus present a potential latency bottleneck.}
\je{The large network weights in these networks also make DRAM I/O a potential bottleneck when storing weights in off-accelerator DRAMs.}

\subsection{\je{Mapping and Dynamic ROIs}}
\je{A final design challenge lies in mapping algorithms running on dynamic ROIs to hardware.}
\je{As established in \cite{ansa}, mapping networks to compute can significantly effect energy efficiency and latency.}
\je{However, processing runtime-dynamic ROIs complicates the generation of these mappings.}
\je{As the range of ROI sizes to be supported is very large, storing a mapping for every possible size is impractical, and therefore generating these mappings entirely offline is imfeasible.}
\je{Simultaneously, the modeling and optimization needed to generate these mappings also prohibits entirely online mapping.}
\je{An intermediate solution is necessary to realize ROI-based stereo depth processing.}

\section{ROI-Based Stereo Depth}\label{sec:algorithm}

\begin{figure}
    \centering
    \includegraphics[width=0.9\linewidth]{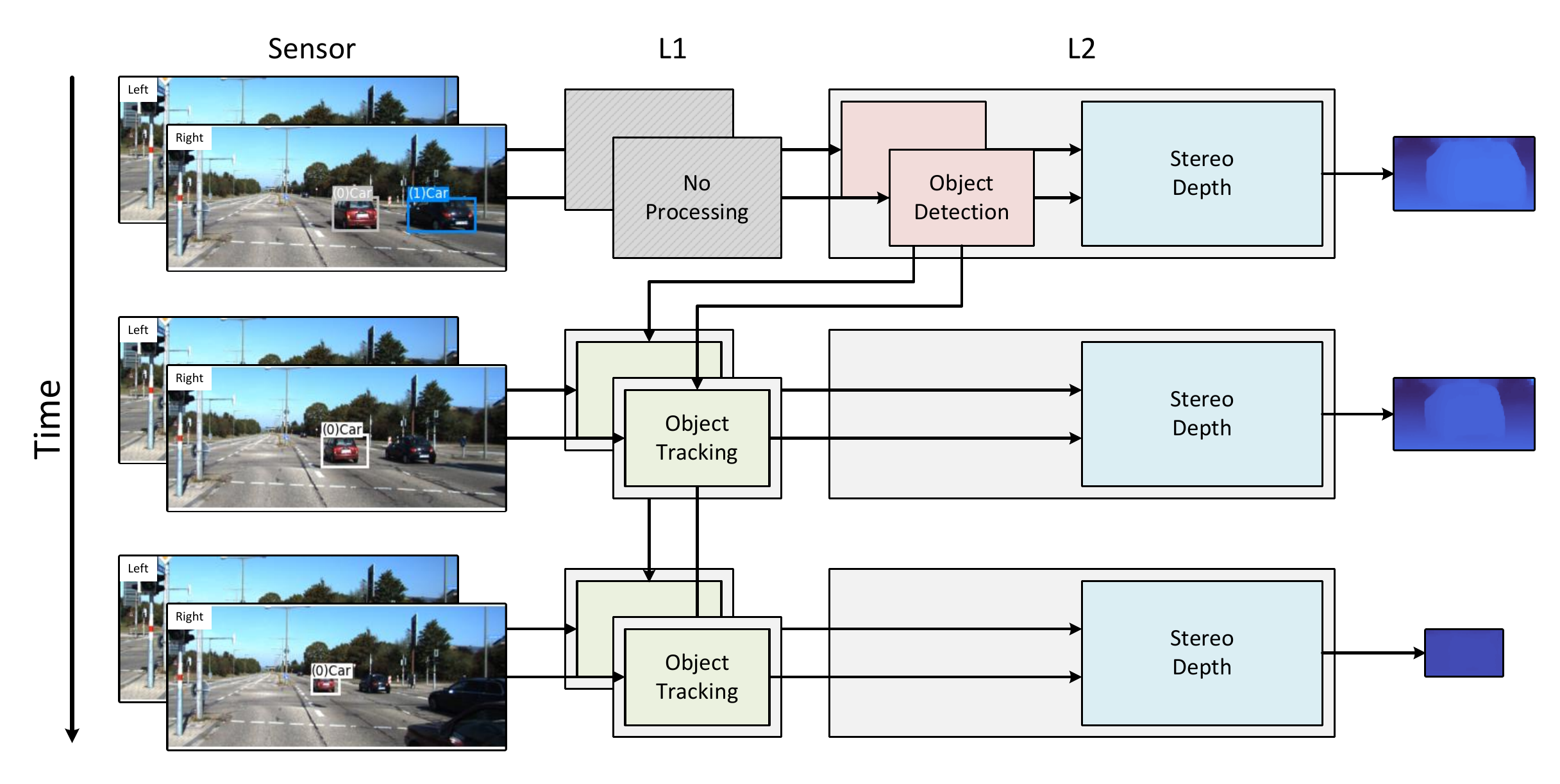}
    \caption{Illustration of the \projname{} processing pipeline. Object detection is run on the L2 processor and run infrequently; object tracking is run on intermediate frames on the L1 processors.}
    \label{fig:alg_design}
\end{figure}

\begin{figure}
    \centering
    \includegraphics[width=0.9\linewidth]{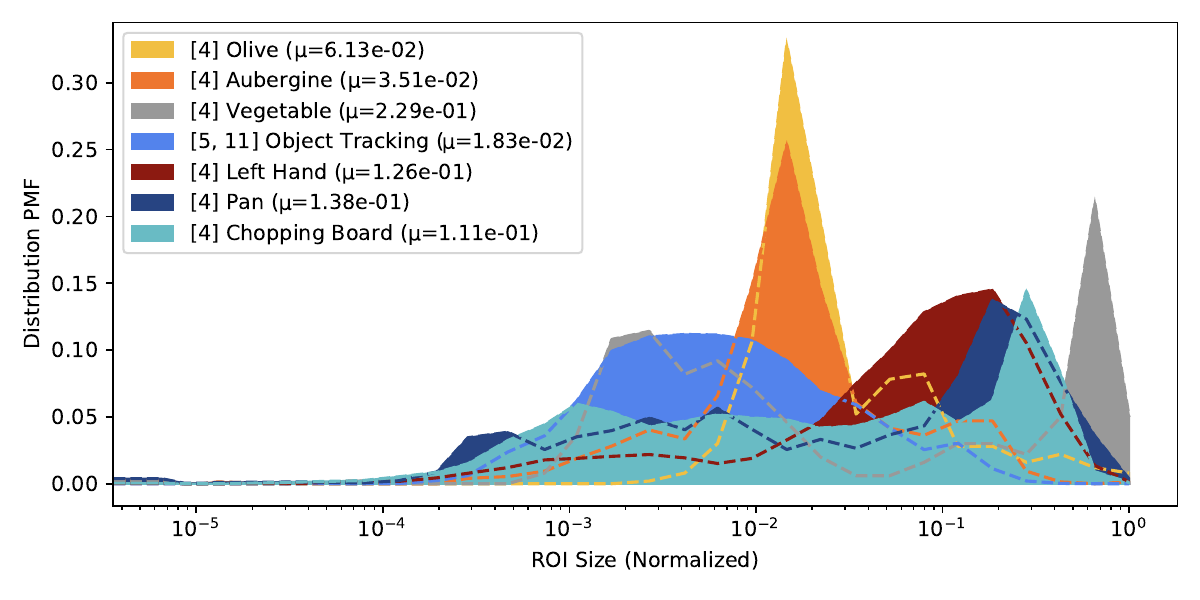}
    \caption{Distribution of ROI sizes across various object tracking datasets and object classes.}
    \label{fig:roi_distribution}
\end{figure}

\je{We propose the use of Regions-of-Interest (ROIs) in AR platforms to augment conventional Stereo Depth Algorithms.}
\je{We illustrate this proposed augmented algorithm in Figure ~\ref{fig:alg_design}.}
In \je{many AR} applications, only specific \je{objects in} an image are of interest\je{; bounding boxes around these objects can be used as ROIs. \cite{eyecod}}
Fig.~\ref{fig:roi_distribution} illustrates the typical sizes of bounding boxes across various object classes in 
\je{egocentric datasets,} KITTI \cite{kitti2012, kitti2015} and Epic Kitchens \cite{epic_kitchens}. 
\je{Typical ROI sizes often multiple orders of magnitude smaller than the full image resolution. }
Unlike classification CNNs, stereo depth models can handle variable-sized ROIs.
This is because they are designed for regression tasks, producing output with the same spatial dimensions as the input images.
With typical ROI sizes often multiple orders of magnitude smaller than the full image resolution, significant processing savings are possible.

\begin{figure}
    \centering
    \includegraphics[width=0.9\linewidth]{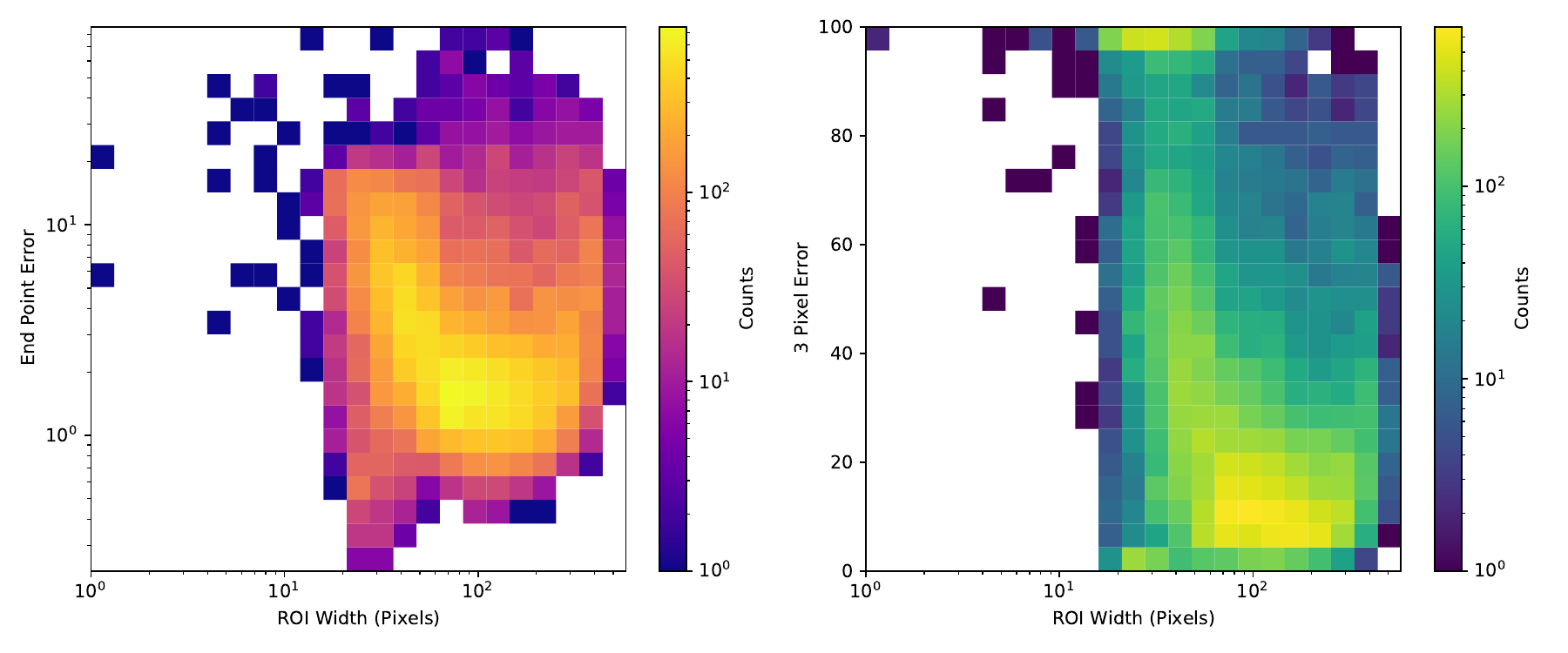}
    \caption{EPE (left) and 3-pixel error (right) for HITNet \cite{hitnet} evaluated on 'Car' ROIs in the KITTI Object Tracking \cite{kitti2012} dataset, as measured against inference on full frames.}
    \label{fig:alg_quality}
\end{figure}

\je{One concern with this method is the degradation of stereo depth quality incurred by processing only ROIs.}
\je{To assess this concern, we evaluate HITNet \cite{hitnet} on crops from the KITTI dataset \cite{kitti2012}, shown in Figure ~\ref{fig:alg_quality}.}
\ca{We choose this dataset for it's availability of full frame stereo depth and object tracking labels.}
\je{We evaluate the degradation of results from full-frame inference by computing End-Point Error (EPE) and 3-pixel error against the full frame results\footnote{Dataset access and model processing took place at the University of Michigan.}; on each graph, lower errors imply less degradation.}
\je{In general, we find that ROI width is most strongly correlated to ROI degradation, with narrow ROIs with little spatial context suffering more compared to broader ROIs.}
\je{However, the tolerability of this degradation has an application dependence.}
\je{While simple algorithms, such as enforcing a minimum ROI size, can be used to address these challenges; in this work, we explore the ramifications of designing a compute system for the full dynamic range of ROI sizes extracted from these datasets.}

While ROIs can reduce stereo depth processing, \je{finding them} can introduce overhead \je{computational} costs.
Extracting ROIs requires object detection \cite{redmon_yolov3_2018}, which demands comparable MACs and weight storage to stereo depth processing itself \cite{hitnet, tiefenrausch}.
Minimizing the overhead of ROI extraction is critical to reduce the loss in efficiency.
\je{We propose to combat this inefficiency by interleaving expensive object detection with fast and efficient object tracking \cite{asu_fpga}, such as a correlation filter \cite{correlation_filter}.}
\je{Such algorithms have been demonstrated to be efficient on low power platforms \cite{vota}.}
\je{For egocentric tasks, where objects move continuously with respect to the observer, such an approach can accurately track objects with greatly reduced computational cost.}
\je{In this work, we specifically use YOLOv3 \cite{redmon_yolov3_2018} and Correlation Filters \cite{correlation_filter} for object detection and tracking, respectively.}
\section{System and Architecture Design}\label{sec:architecture}

\begin{figure*}
    \centering
    \includegraphics[width=0.8\linewidth]{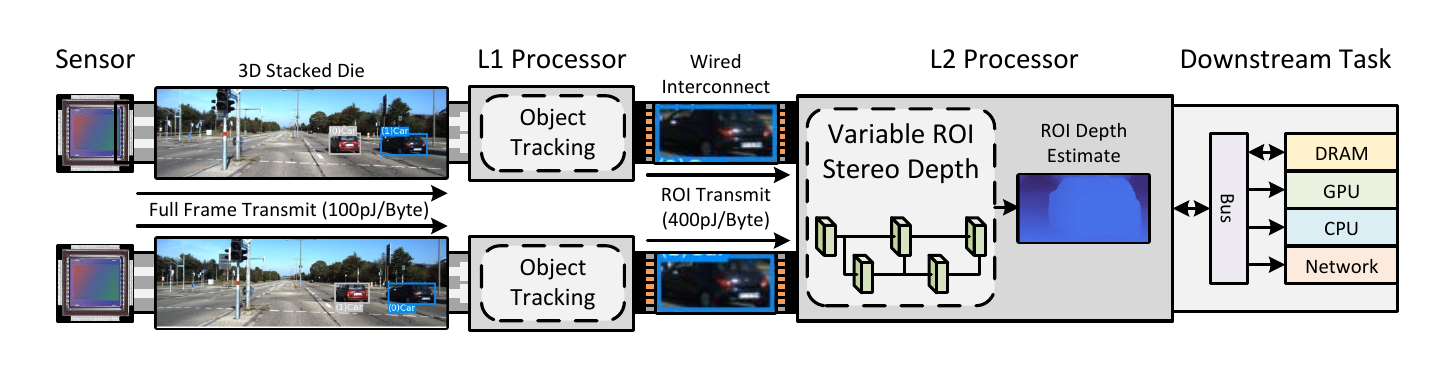}
    \caption{\projname{} system design.}
    \label{fig:steroi_compare}
\end{figure*}

\textbf{System Design}:
The system-level design of \projname{} is depicted in Fig.~\ref{fig:steroi_compare}. 
\je{The design features stereo sensors and a custom accelerator integrated into an SoC, which is a typical design for AR platforms \cite{aria}.}
\je{Furthermore, each sensor is co-packaged with a lightweight L1 processor, based on \cite{siracusa}}.
\je{These processors are responsible for handling the object tracking algorithms used to extract ROIs; in this way, they enable saving energy when transmitting ROIs from the L1 processor to the SoC and the accelerator (heretoafter referred to as the \textit{L2 Processor}.}

\begin{figure}
    \centering
    \includegraphics[width=0.8\linewidth]{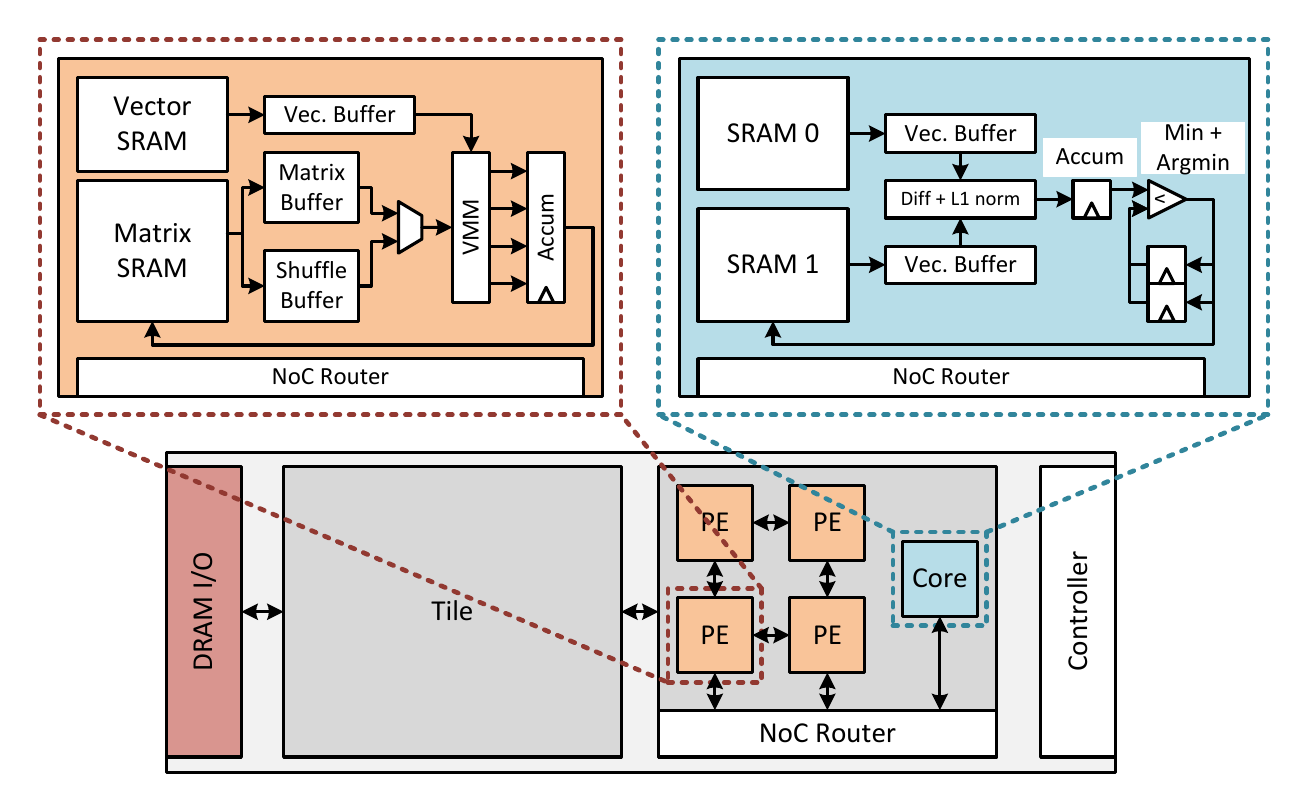}
    \caption{\projname{} L2 processor architecture. The Special Compute Unit (SCU) accelerates non-parameterized compute patterns.}
    \label{fig:arch_sketch}
\end{figure}

\je{\textbf{Accelerator Architecture}:}
The L2 processor is responsible for object detection and ROI-based stereo depth processing 
The \je{accelerator} uses a parameterized hierarchical \je{architecture}, pictured in Fig.~\ref{fig:arch_sketch}, which draws inspiration from ANSA \cite{ansa}.
The architecture is composed of tiles, with each tile consisting of a collection of PEs. 
Each PE is responsible for executing convolutions and activation functions through a Vector-Matrix Multiplier (VMM), local SRAMs, and a shuffle buffer for depthwise convolutions.
Flexibility is achieved through both the hierarchical structure, allowing for dynamic reconfiguration of compute resources for different compute tasks\je{;} and the compute units themselves, which support multiple dataflows to enable optimization at the mapping level.
\je{Power gating is further used to capture further energy savings by disabling unused resources on a per-frame basis.}
We have expanded on ANSA to accommodate the diverse sizes of ROIs while maintaining energy efficiency. 

\begin{figure}
    \centering
    \includegraphics[width=0.75\linewidth]{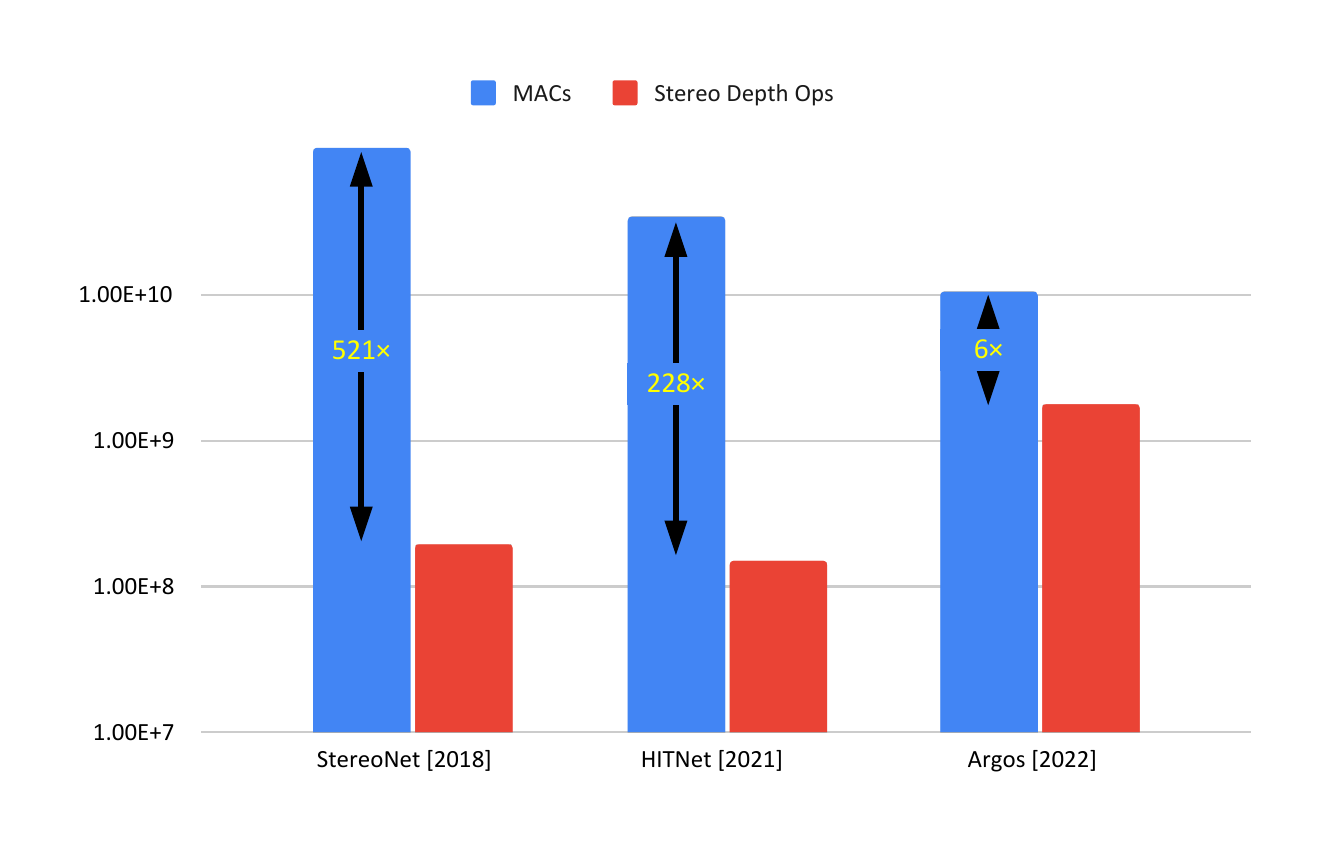}
    \caption{Conventional CNN operation counts (e.g. convolution MACs) and stereo depth specific operations in stereo depth networks in recent years on 384$\times$1280 images.}
    \label{fig:ops_compare}
\end{figure}

\textbf{Special Compute Unit (SCU)}: 
Stereo depth networks consist of both CNN layers and stereo-depth-specific non-parameterized layers, as illustrated in Fig.~\ref{fig:ops_compare}.
These operations, seen in many networks \cite{stereonet, hitnet, tiefenrausch}, are used for processing disparity estimates.
\je{They use operations and data broadcast patterns for which conventional linear algebra engines are ill-suited, such as vector L1 norm and list argmin.}
\je{While not the majority of operations in any network, they constitute a sufficient portion of the network to necessitate hardware support to enable low latency, high framerate processing.}
Supporting these operations directly in PEs would increase their complexity and footprint, and reduce efficiency for both CNN and special operations.
Therefore, the \projname{} L2 processor uses both PEs for CNN compute, and special compute units (SCUs) for special operations.

To design an efficient SCU, we first observe that the cost volume processing in \cite{stereonet, hitnet, tiefenrausch} employs a limited set of compute primitives: vector-vector difference, L1 norm, sequence minimum, and argmin.
We propose a SCU pipeline with bypass options to handle these operations as well as their compositions.
The resulting SCU design captures various variants of cost volume processing for each stereo depth network, along with the warp and aggregate operations from \cite{hitnet}, and the maxpool operation in \cite{redmon_yolov3_2018}. 
\ca{A single SCU is allocated per Tile; the correct balance of SCUs to PEs is then realized through design space exploration over the mapping design.}

\textbf{Mutipacket NoC Routing}: 
The \projname{} L2 processor uses hierarchically arranged mesh NoCs connecting tiles globally and PEs locally.
These NoCs allow for fine-grained partitioning of compute tasks. 
However, this flexibility also costs redundant data movements.
To mitigate this cost, the \projname{} L2 processor uses a multipacket NoC. 
A multipacket consists of a data packet and a list of destination nodes. 
When a NoC node receives a multipacket, it forwards the data to the remaining destination nodes in the network.
To minimize the overhead of this scheme, we use simple Direction Order Routing (DOR) \cite{eecs570_dor}.
With DOR, each data packet is sent over a given link only once.
This approach reduces data movement while maintaining high flexibility in compute unit allocation.

\section{\projname{} L2 Processor Mapping}\label{sec:mapping}


\je{To leverage the energy savings made possible by our algorithm and architecture, we must do a good job of mapping compute onto the processor.}
\je{However, there are two key challenges that must be overcome to achieve this.}
\je{The first is the vast space of possible mappings that are possible; to find low energy, low latency mappings within this space, we must either reduce the dimensionality of this space, or develop algorithms to efficiently traverse it.}
\je{The variable ROI-size in this application adds the second challenge of providing mapping support across this range of possible ROIs.}
\je{Purely offline solutions to this problem are impractical, due to the storage requirements for supporting the full ROI range; and purely online solutions are equally impractical, due to the complexity of generating these mappings.}

\subsection{Single-ROI Mapping}\label{subsec:single_roi}

\je{We first consider the space of mappings for algorithms running on a fixed ROI size.}
\je{To traverse the large space of possible mappings, we first generate a set of higher level mapping descriptors, which can be used to generate low-level control signals for the processor.}

\textbf{DRAM I/O Options}:
\je{The off-chip DRAM storage in the \projname{} system can be used to store intermediate activations in neural networks.}
\je{This helps to address two levels of memory utilization variance.}
\je{Within a single frame, different activations in the network can have vastly different sizes; to enable processing on form-factor limited processors, it is beneficial to not rely on local SRAMs to store these activations.}
\je{Across multiple frames, different ROI sizes result in similarly variable activation sizes.}
DRAM provides a means to trade off dynamic energy and latency with SRAM utilization for handling these large ROIs and layers: firstly, by choosing which activations are stored in DRAM; and secondly, by choosing how these activations are accessed.
Activations may be streamed directly from DRAM to local buffers to reduce SRAM utilization. 
Alternatively, activations can be buffered in SRAM to reduce energy and latency.

\begin{figure}
    \centering
    \includegraphics[width=0.75\linewidth]{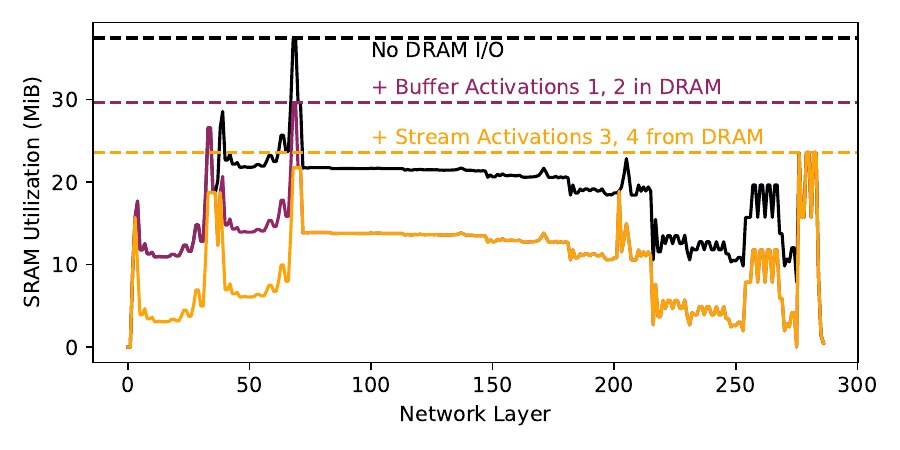}
    \caption{By successively buffering or streaming large activations from DRAM, a progression of \textit{DRAM Modes} are formed which reduce SRAM utilization.}
    \label{fig:dram_modes}
\end{figure}

\je{To choose which activations to store in DRAM and which to store in SRAM, we consider the minimal amount of off-device storage necessary to reduce an algorithms peak SRAM utilization, as seen in Figure ~\ref{fig:dram_modes}.}
\je{Iterating on this process, we generate a sequence of \textit{DRAM Modes}, or sets of activations to be streamed or buffered in DRAM.}
\je{By forming this sequence, we reduce the problem of determining DRAM I/O for a mapping from making a per-activation ternary choice, to selecting selecting the appropriate DRAM mode to fit within a given architecture's memory budget.}
\je{In practice, a combination of dataflow selection and DRAM I/O can be used to reduce SRAM utilization; therefore, we generate mappings for a few DRAM modes, and evaluate each on latency, energy, and memory utilization when generating mappings.}

\textbf{Dataflow Options}: The \projname{} L2 processor supports per-layer dataflows for both PEs and SCUs.
These dataflows include Weight and Input Stationary, which define the patterns of data reuse for weights and activations\je{;} and Channel First and Last, which divide input channel accumulation either spatially or temporally.
While the choice between Weight and Input Stationary is conventionally based on data reuse, the relative sizes of activations and weights for variably sized ROIs must also be considered.
Data movement in the NoC must also be managed. Different dataflows affect the location where activations are produced, thereby impacting the efficiency of data movement between PEs.

\je{Prior works \cite{ansa} have used greedy algorithms to assign per-layer dataflows to a network.}
\je{While this method is useful for minimizing latency and dynamic energy; we find it is insufficient to satisfy SRAM utilization constraints, as minimizing peak memory utilization requires global optimization.}
\je{Therefore, we use an augmented greedy algorithm to assign per-layer dataflows, which consists of a preliminary greedy assignment to minimize dynamic energy, and a subsequent optimization step which identifies dataflow choices that exceed memory useage limits.}

\je{\textbf{Tile Shutoff}:}
\je{We also consider turning off tiles, PEs, or SCUs within a processor to be an aspect of mapping.}
\je{Partial processor shutoff can be used on a per-frame basis to conform the compute capacity of the processor to the current task.}
\je{This allows dynamic energy efficiency to be sacrificed in order to reduce static power draw for the duration of processing a given frame.}
\je{Like with DRAM Modes, we limit our evaluation to a small number of processor sub-configurations to simplify the mapping design space.}

\subsection{Multiple-ROI Binning}\label{subsec:multi_roi}

\je{We consider the combination of a DRAM Mode, a dataflow assignment for each network layer, and a processor sub-configuration to constitute a \textit{mapping descriptor}.}
\je{From this mapping descriptor, a \textit{low level mapping} may be trivially generated by assigning compute operations and activation storage to the PEs, SCUs, and SRAMs within a processor.}
\je{While generating a low-level mapping requires knowledge of the ROI and intermediate feature sizes, a mapping descriptor is agnostic of this detail.}
\je{For HITNet, storing a mapping descriptor takes on the order of 100s of Bytes of memory.}

\je{To support the full range of possible ROI sizes efficiently, we propose to divide this range of sizes into a finite number of intervals, and assigning a separate high level mapping descriptor to each interval.}
\je{Low level mapping can then be performed at runtime with minimal overhead.}
\je{This scheme allows for the mapping descriptors to be optimized offline while still enabling runtime ROI variability.}
\je{Additionally, using separate mapping descriptors for different ROI size intervals allows for mappings to be customized for each size; for example, larger ROIs may use mappings that focus more heavily on reducing SRAM utilization, while mappings for smaller ROIs can focus entirely on maximizing energy efficiency.}
\je{We combine optimized mapping design with architecture architectural design exploration.}
\je{In this way, we are able to determine the optimal architecture design for different ROI probability distributions.}

\section{Results}\label{sec:results}

We focus our analysis on the L2 processor design and performance.
For the evaluations, HITNet \cite{hitnet} has been chosen as a representative of the latest stereo depth algorithms.
For comparison with the L2 processor, we evaluate HITNet on the Jetson Orin Nano, which is a representative off-the-shelf mobile compute system.
A public implementation of HITNet is used \cite{tinyhitnet}, and it is compiled on a per-ROI size basis using ONNX and TensorRT. This approach provides optimistic estimates for the device's performance.
To measure system power, we utilize Jetson Stats and isolate the power consumption of the GPU and CPU for comparison.

Our system simulator is made of multiple parts. 
Firstly, we implement a counter-based architecture model to estimate the L2 processor performance.
The dynamic and static energy of PE, SCU, \ca{and} SRAM I/O are estimated based on post-APR simulation using the TSMC 28nm PDK and 16-bit operations.
This L2 simulator also accounts for DRAM I/O \cite{lpddr4x, lpddr5_est} and NoC \cite{ansa} energy and latency.
We base the L1 energy and latency on \cite{marsellus} and estimate the energy required for object detection on images of size $384\times1280$.
We also evaluate system level sensor \cite{gs_cis1}, uTSV, and MIPI interface \cite{gomez_distributed_2022} energy.
This complete simulator is used to conduct design space sweeps of the \projname{} system running HITNet for stereo depth and TinyYOLOv3 for object detection, according to Section ~\ref{sec:mapping}.

\subsection{Ablation Studies}



\begin{figure}
    \centering
    \includegraphics[width=0.8\linewidth]{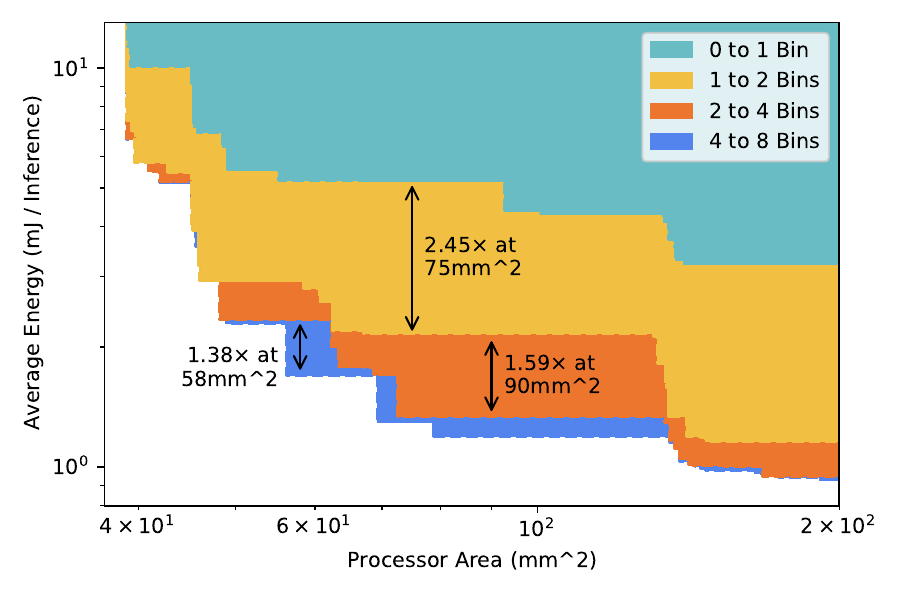}
    \caption{Effect of number of bins. Increasing bin count results in marginal gains, with highest energy savings on intermediate sized processors.}
    \label{fig:result_bincount_ablation}
\end{figure}

\textbf{Bin Count}: In Fig.~\ref{fig:result_bincount_ablation}, we evaluate the results using different number of bins, which corresponds to the number of runtime ROI intervals.
Moving from one to two bins, the processor can use different mappings for large and small ROI sizes, optimized for different objectives.
This results in a significant gain in energy.
However, as the number of bins increases beyond 2 bins, the marginal gain per additional bin diminishes and varies slightly across processor areas.
In some cases, adding an extra bin can still be beneficial to better adapt to the specific ROI distribution being used.

\begin{figure}
    \centering
    \includegraphics[width=0.8\linewidth]{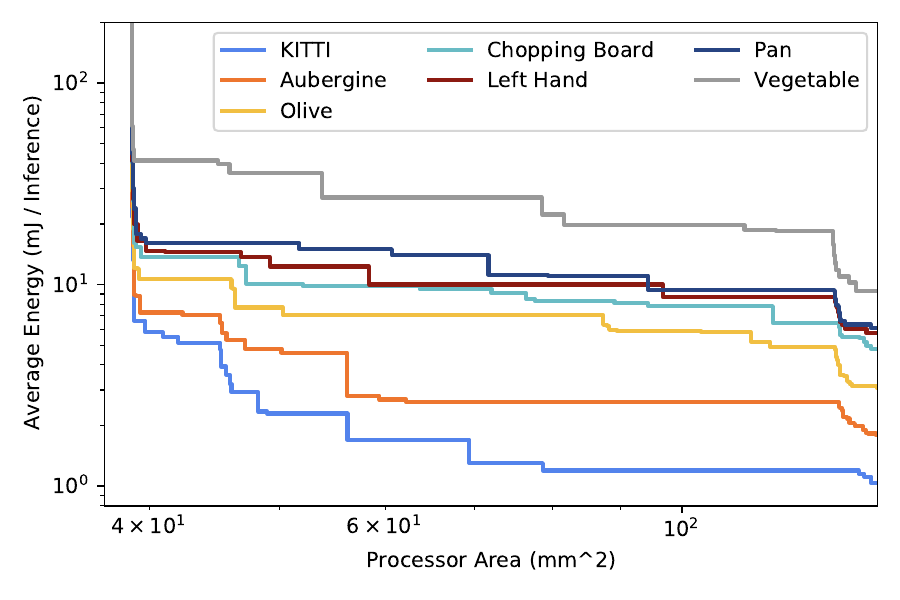}
    \caption{Results of different ROI distributions. The pareto curve of an ROI distribution is dictated primarily by the ROI mean, though variance also plays a minor role.}
    \label{fig:result_distribution_compare}
\end{figure}

\textbf{ROI Distributions}: We compare results for multiple ROI probability distributions to evaluate the generality of our design methodology, as seen in Fig.~\ref{fig:result_distribution_compare}.
Interestingly, the average energy required to run the various ROI distributions is ordered in the same sequence as their mean ROI size, as reported in Fig.~\ref{fig:roi_distribution}.
This suggests that our design method minimizes nonlinear overheads caused by the variable ROI sizes.
Even high variance bimodal distributions, such as ``Vegetables'' \cite{epic_kitchens}, can be efficiently handled by adequately parameterized ROI binnings on \projname{}.

\subsection{Design Benchmarking}

Next, we compare designs generated by this methodology, with existing edge system and baseline ASIC designs.

\begin{figure}
    \centering
    \includegraphics[width=0.8\linewidth]{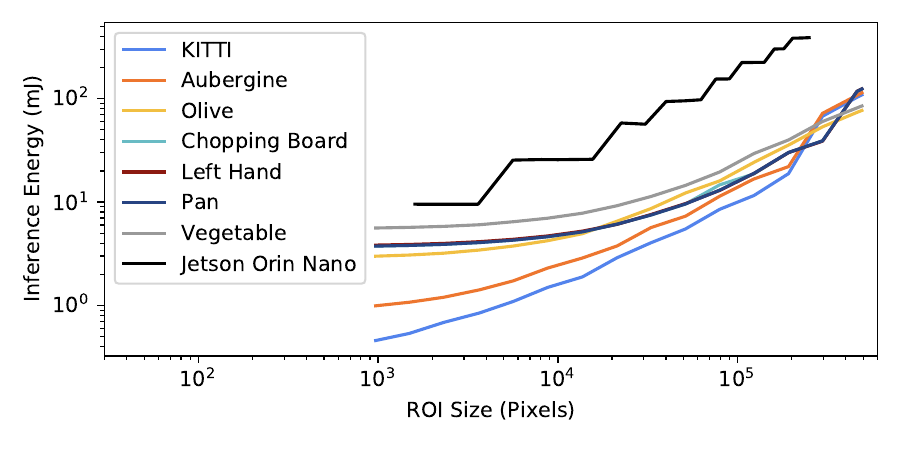}
    \caption{Energy consumption of Jetson Orin Nano running different ROI sizes and \projname{} systems optimized for different ROI distributions. \projname{} achieves superior granularity in energy and lower overall energy.}
    \label{fig:jetson_compare}
\end{figure}

\textbf{Comparison with Jetson Orin Nano}: 
In Fig.~\ref{fig:jetson_compare}, we compare the per-ROI energy of the Jetson Orin Nano with \projname{} systems optimized for different ROI distributions.
The \projname{} designs demonstrate two prominent advantages.
Firstly, a \projname{} design can be optimized based on the ROI distribution, whereas the Jetson Orin Nano requires statically compiled binaries for each ROI size in the distribution, which makes it less practical. 
Secondly, 
\projname{} processor dynamically scales performance and energy according to the size of ROI being processed.
In contrast, the Jetson Orin Nano appears to suffer from a coarse-grained reconfigurability of its tensor cores, resulting in an energy pattern characterized by stair steps\ca{; compute latency also suffers, and the Jetson Orin Nano does not exceed 15 FPS operation.}
\projname{}, on the other hand, uses tiles, PEs and SCUs to enable fine-grained optimization.

\begin{figure}
    \centering
    \includegraphics[width=0.9\linewidth]{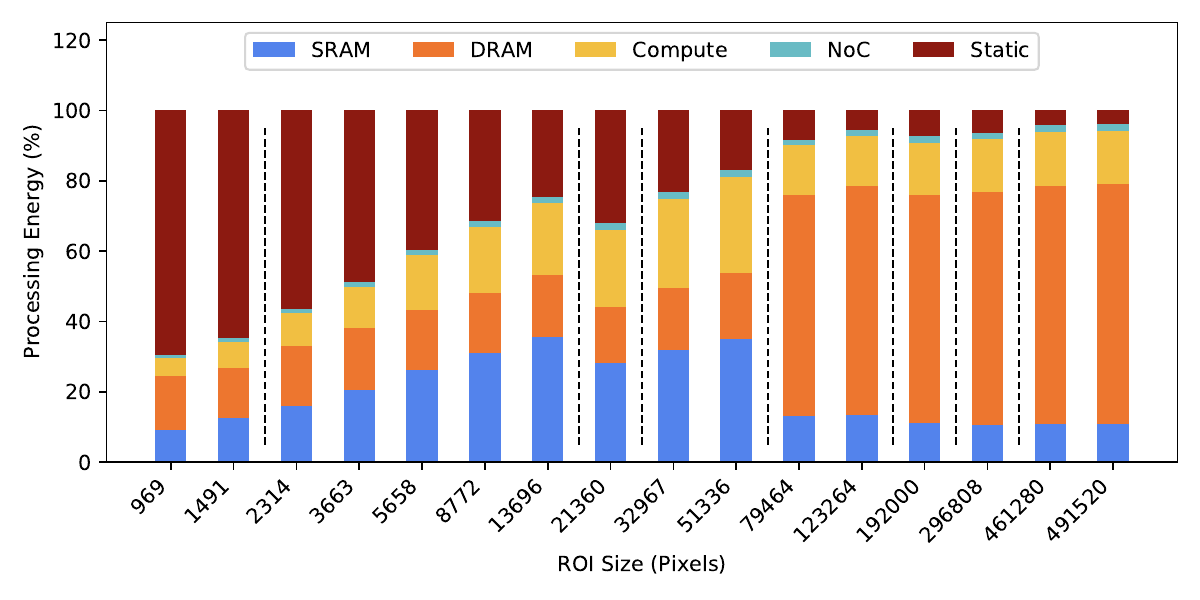}
    \caption{Breakdown of \projname{} L2 processor energy by ROI size. Dashed lines represent the boundaries of mapping bins.}
    \label{fig:binning_breakdown}
\end{figure}

\textbf{Energy Breakdown by ROI Size}: We analyze the breakdown of energy by ROI size in a \projname{} L2 processor, as illustrated in Fig.~\ref{fig:binning_breakdown}.
The energy consumption is primarily influenced by static power and DRAM access, with their proportions varying accordingly.
For very small ROIs, static power draw is dominant. 
For extremely large ROIs, the energy is dominated by DRAM I/O.
DRAM I/O is used to minimize the L2 SRAM and reduce static power for small ROIs.
This insight underscores the challenge of optimizing energy across ROI distribution in the L2 processor design. 

\begin{figure}
    \centering
    \includegraphics[width=0.9\linewidth]{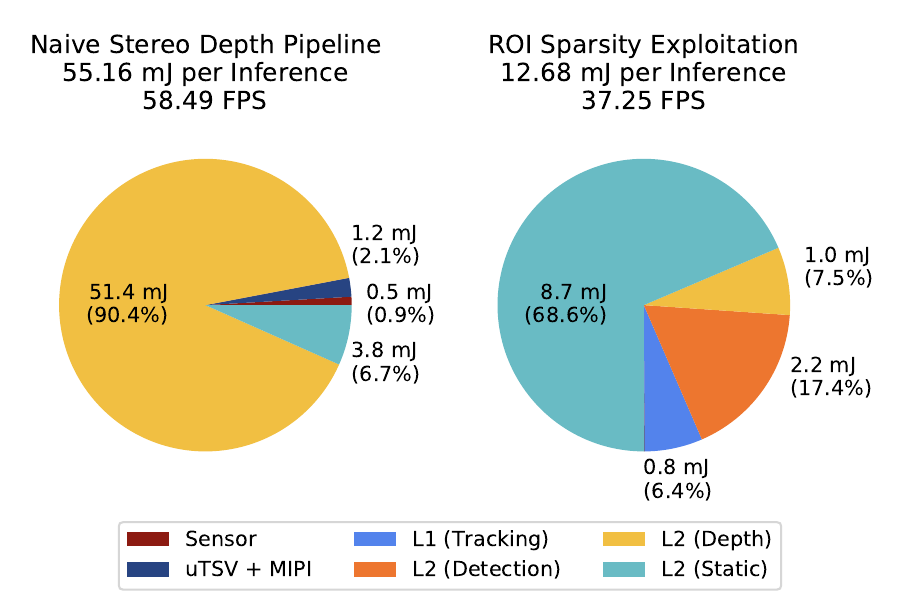}
    \caption{Comparison of energy in baseline design (no ROI exploitation) with \projname{} design, assuming object detection runs every 5 frames and KITTI ROI distribution.}
    \label{fig:baseline_compare}
\end{figure}

\textbf{Comparison with Baseline System}: In Fig.~\ref{fig:baseline_compare}, we compare a baseline system with no ROI or temporal sparsity with a \projname{} system.
In this comparison, we optimize both processors to have an area under 100 mm\textsuperscript{2} and a frame rate of at least 30 FPS.
The exploitation of ROI requires object detection and object tracking.
Energy savings are limited by these two costs, which do not scale with the ROI.
Nevertheless, \projname{} still achieves \textbf{$4.35\times$} per-inference energy savings by effectively leveraging ROI-based processing.

\section{Conclusion}\label{sec:conclusion}

In this work, we presented \textit{\projname{}}, a novel system design and mapping framework aimed at achieving energy-efficient stereo depth processing in AR/VR devices.
By considering the distinct compute requirements on image regions of multiple orders-of-magnitude of scales, our mapping and system co-design loop produces processor designs that fully realize the potential energy savings of ROI-based processing.
Additionally, we propose several specialized compute primitives that enhance efficiency for the unique compute patterns found in stereo depth processing.
Our evaluations demonstrate superior scalability compared to a conventional edge compute platform, and up to $4.35\times$ energy savings over a baseline non-ROI-based processing systems.

\begin{acks}
This project was funded by the NDSEG Fellowship, and by Meta Reality Labs Research.
\end{acks}

\bibliographystyle{ACM-Reference-Format}
\bibliography{refs}


\begin{thebibliography}{29}


\ifx \showCODEN    \undefined \def \showCODEN     #1{\unskip}     \fi
\ifx \showDOI      \undefined \def \showDOI       #1{#1}\fi
\ifx \showISBNx    \undefined \def \showISBNx     #1{\unskip}     \fi
\ifx \showISBNxiii \undefined \def \showISBNxiii  #1{\unskip}     \fi
\ifx \showISSN     \undefined \def \showISSN      #1{\unskip}     \fi
\ifx \showLCCN     \undefined \def \showLCCN      #1{\unskip}     \fi
\ifx \shownote     \undefined \def \shownote      #1{#1}          \fi
\ifx \showarticletitle \undefined \def \showarticletitle #1{#1}   \fi
\ifx \showURL      \undefined \def \showURL       {\relax}        \fi
\providecommand\bibfield[2]{#2}
\providecommand\bibinfo[2]{#2}
\providecommand\natexlab[1]{#1}
\providecommand\showeprint[2][]{arXiv:#2}

\bibitem[qua({[n.\,d.]})]%
        {qualcomm_soc}
 \bibinfo{year}{[n.\,d.]}\natexlab{}.
\newblock \bibinfo{title}{qualcomm.com}.
\newblock \bibinfo{howpublished}{\url{https://www.qualcomm.com/content/dam/qcomm-martech/dm-assets/documents/whitepaper_-_driving_the_new_era_of_immersive_experiences_-_qualcomm.pdf}}.
\newblock
\newblock
\shownote{[Accessed 20-11-2024]}.


\bibitem[Bolme et~al\mbox{.}(2010)]%
        {correlation_filter}
\bibfield{author}{\bibinfo{person}{David~S. Bolme}, \bibinfo{person}{J.~Ross Beveridge}, \bibinfo{person}{Bruce~A. Draper}, {and} \bibinfo{person}{Yui~Man Lui}.} \bibinfo{year}{2010}\natexlab{}.
\newblock \showarticletitle{Visual object tracking using adaptive correlation filters}. In \bibinfo{booktitle}{\emph{2010 {IEEE} {Computer} {Society} {Conference} on {Computer} {Vision} and {Pattern} {Recognition}}}. \bibinfo{pages}{2544--2550}.
\newblock
\urldef\tempurl%
\url{https://doi.org/10.1109/CVPR.2010.5539960}
\showDOI{\tempurl}
\newblock
\shownote{ISSN: 1063-6919}.


\bibitem[Choi et~al\mbox{.}(2018)]%
        {depth_gesture_recog}
\bibfield{author}{\bibinfo{person}{Sungpill Choi}, \bibinfo{person}{Jinsu Lee}, \bibinfo{person}{Kyuho Lee}, {and} \bibinfo{person}{Hoi-Jun Yoo}.} \bibinfo{year}{2018}\natexlab{}.
\newblock \showarticletitle{A 9.02mW CNN-stereo-based real-time 3D hand-gesture recognition processor for smart mobile devices}. In \bibinfo{booktitle}{\emph{2018 IEEE International Solid-State Circuits Conference - (ISSCC)}}. \bibinfo{pages}{220--222}.
\newblock
\urldef\tempurl%
\url{https://doi.org/10.1109/ISSCC.2018.8310263}
\showDOI{\tempurl}


\bibitem[Conti et~al\mbox{.}(2024)]%
        {marsellus}
\bibfield{author}{\bibinfo{person}{Francesco Conti}, \bibinfo{person}{Gianna Paulin}, \bibinfo{person}{Angelo Garofalo}, \bibinfo{person}{Davide Rossi}, \bibinfo{person}{Alfio Di~Mauro}, \bibinfo{person}{Georg Rutishauser}, \bibinfo{person}{Gianmarco Ottavi}, \bibinfo{person}{Manuel Eggimann}, \bibinfo{person}{Hayate Okuhara}, {and} \bibinfo{person}{Luca Benini}.} \bibinfo{year}{2024}\natexlab{}.
\newblock \showarticletitle{Marsellus: {A} {Heterogeneous} {RISC}-{V} {AI}-{IoT} {End}-{Node} {SoC} with 2-to-8b {DNN} {Acceleration} and 30\%-{Boost} {Adaptive} {Body} {Biasing}}.
\newblock \bibinfo{journal}{\emph{IEEE Journal of Solid-State Circuits}} \bibinfo{volume}{59}, \bibinfo{number}{1} (\bibinfo{date}{Jan.} \bibinfo{year}{2024}), \bibinfo{pages}{128--142}.
\newblock
\showISSN{0018-9200, 1558-173X}
\urldef\tempurl%
\url{https://doi.org/10.1109/JSSC.2023.3318301}
\showDOI{\tempurl}
\newblock
\shownote{arXiv:2305.08415 [cs]}.


\bibitem[Damen et~al\mbox{.}(2022)]%
        {epic_kitchens}
\bibfield{author}{\bibinfo{person}{Dima Damen}, \bibinfo{person}{Hazel Doughty}, \bibinfo{person}{Giovanni~Maria Farinella}, \bibinfo{person}{Antonino Furnari}, \bibinfo{person}{Jian Ma}, \bibinfo{person}{Evangelos Kazakos}, \bibinfo{person}{Davide Moltisanti}, \bibinfo{person}{Jonathan Munro}, \bibinfo{person}{Toby Perrett}, \bibinfo{person}{Will Price}, {and} \bibinfo{person}{Michael Wray}.} \bibinfo{year}{2022}\natexlab{}.
\newblock \showarticletitle{Rescaling Egocentric Vision: Collection, Pipeline and Challenges for EPIC-KITCHENS-100}.
\newblock \bibinfo{journal}{\emph{International Journal of Computer Vision (IJCV)}}  \bibinfo{volume}{130} (\bibinfo{year}{2022}), \bibinfo{pages}{33–55}.
\newblock
\urldef\tempurl%
\url{https://doi.org/10.1007/s11263-021-01531-2}
\showURL{%
\tempurl}


\bibitem[Eki et~al\mbox{.}(2021)]%
        {sony_coprocessor}
\bibfield{author}{\bibinfo{person}{Ryoji Eki}, \bibinfo{person}{Satoshi Yamada}, \bibinfo{person}{Hiroyuki Ozawa}, \bibinfo{person}{Hitoshi Kai}, \bibinfo{person}{Kazuyuki Okuike}, \bibinfo{person}{Hareesh Gowtham}, \bibinfo{person}{Hidetomo Nakanishi}, \bibinfo{person}{Edan Almog}, \bibinfo{person}{Yoel Livne}, \bibinfo{person}{Gadi Yuval}, \bibinfo{person}{Eli Zyss}, {and} \bibinfo{person}{Takashi Izawa}.} \bibinfo{year}{2021}\natexlab{}.
\newblock \showarticletitle{9.6 A 1/2.3inch 12.3Mpixel with On-Chip 4.97TOPS/W CNN Processor Back-Illuminated Stacked CMOS Image Sensor}. In \bibinfo{booktitle}{\emph{2021 IEEE International Solid-State Circuits Conference (ISSCC)}}, Vol.~\bibinfo{volume}{64}. \bibinfo{pages}{154--156}.
\newblock
\urldef\tempurl%
\url{https://doi.org/10.1109/ISSCC42613.2021.9365965}
\showDOI{\tempurl}


\bibitem[Engel et~al\mbox{.}(2023)]%
        {aria}
\bibfield{author}{\bibinfo{person}{Jakob Engel}, \bibinfo{person}{Kiran Somasundaram}, \bibinfo{person}{Michael Goesele}, \bibinfo{person}{Albert Sun}, \bibinfo{person}{Alexander Gamino}, \bibinfo{person}{Andrew Turner}, \bibinfo{person}{Arjang Talattof}, \bibinfo{person}{Arnie Yuan}, \bibinfo{person}{Bilal Souti}, \bibinfo{person}{Brighid Meredith}, \bibinfo{person}{Cheng Peng}, \bibinfo{person}{Chris Sweeney}, \bibinfo{person}{Cole Wilson}, \bibinfo{person}{Dan Barnes}, \bibinfo{person}{Daniel DeTone}, \bibinfo{person}{David Caruso}, \bibinfo{person}{Derek Valleroy}, \bibinfo{person}{Dinesh Ginjupalli}, \bibinfo{person}{Duncan Frost}, \bibinfo{person}{Edward Miller}, \bibinfo{person}{Elias Mueggler}, \bibinfo{person}{Evgeniy Oleinik}, \bibinfo{person}{Fan Zhang}, \bibinfo{person}{Guruprasad Somasundaram}, \bibinfo{person}{Gustavo Solaira}, \bibinfo{person}{Harry Lanaras}, \bibinfo{person}{Henry Howard-Jenkins}, \bibinfo{person}{Huixuan Tang}, \bibinfo{person}{Hyo~Jin Kim}, \bibinfo{person}{Jaime Rivera},
  \bibinfo{person}{Ji Luo}, \bibinfo{person}{Jing Dong}, \bibinfo{person}{Julian Straub}, \bibinfo{person}{Kevin Bailey}, \bibinfo{person}{Kevin Eckenhoff}, \bibinfo{person}{Lingni Ma}, \bibinfo{person}{Luis Pesqueira}, \bibinfo{person}{Mark Schwesinger}, \bibinfo{person}{Maurizio Monge}, \bibinfo{person}{Nan Yang}, \bibinfo{person}{Nick Charron}, \bibinfo{person}{Nikhil Raina}, \bibinfo{person}{Omkar Parkhi}, \bibinfo{person}{Peter Borschowa}, \bibinfo{person}{Pierre Moulon}, \bibinfo{person}{Prince Gupta}, \bibinfo{person}{Raul Mur-Artal}, \bibinfo{person}{Robbie Pennington}, \bibinfo{person}{Sachin Kulkarni}, \bibinfo{person}{Sagar Miglani}, \bibinfo{person}{Santosh Gondi}, \bibinfo{person}{Saransh Solanki}, \bibinfo{person}{Sean Diener}, \bibinfo{person}{Shangyi Cheng}, \bibinfo{person}{Simon Green}, \bibinfo{person}{Steve Saarinen}, \bibinfo{person}{Suvam Patra}, \bibinfo{person}{Tassos Mourikis}, \bibinfo{person}{Thomas Whelan}, \bibinfo{person}{Tripti Singh}, \bibinfo{person}{Vasileios Balntas},
  \bibinfo{person}{Vijay Baiyya}, \bibinfo{person}{Wilson Dreewes}, \bibinfo{person}{Xiaqing Pan}, \bibinfo{person}{Yang Lou}, \bibinfo{person}{Yipu Zhao}, \bibinfo{person}{Yusuf Mansour}, \bibinfo{person}{Yuyang Zou}, \bibinfo{person}{Zhaoyang Lv}, \bibinfo{person}{Zijian Wang}, \bibinfo{person}{Mingfei Yan}, \bibinfo{person}{Carl Ren}, \bibinfo{person}{Renzo~De Nardi}, {and} \bibinfo{person}{Richard Newcombe}.} \bibinfo{year}{2023}\natexlab{}.
\newblock \bibinfo{title}{Project Aria: A New Tool for Egocentric Multi-Modal AI Research}.
\newblock
\newblock
\showeprint[arxiv]{2308.13561}~[cs.HC]
\urldef\tempurl%
\url{https://arxiv.org/abs/2308.13561}
\showURL{%
\tempurl}


\bibitem[Geiger et~al\mbox{.}(2012)]%
        {kitti2012}
\bibfield{author}{\bibinfo{person}{Andreas Geiger}, \bibinfo{person}{Philip Lenz}, {and} \bibinfo{person}{Raquel Urtasun}.} \bibinfo{year}{2012}\natexlab{}.
\newblock \showarticletitle{Are we ready for Autonomous Driving? The KITTI Vision Benchmark Suite}. In \bibinfo{booktitle}{\emph{Conference on Computer Vision and Pattern Recognition (CVPR)}}.
\newblock


\bibitem[Gomez et~al\mbox{.}(2022)]%
        {gomez_distributed_2022}
\bibfield{author}{\bibinfo{person}{Jorge Gomez}, \bibinfo{person}{Saavan Patel}, \bibinfo{person}{Syed~Shakib Sarwar}, \bibinfo{person}{Ziyun Li}, \bibinfo{person}{Raffaele Capoccia}, \bibinfo{person}{Zhao Wang}, \bibinfo{person}{Reid Pinkham}, \bibinfo{person}{Andrew Berkovich}, \bibinfo{person}{Tsung-Hsun Tsai}, \bibinfo{person}{Barbara De~Salvo}, {and} \bibinfo{person}{Chiao Liu}.} \bibinfo{year}{2022}\natexlab{}.
\newblock \bibinfo{title}{Distributed {On}-{Sensor} {Compute} {System} for {AR}/{VR} {Devices}: {A} {Semi}-{Analytical} {Simulation} {Framework} for {Power} {Estimation}}.
\newblock
\newblock
\urldef\tempurl%
\url{https://doi.org/10.48550/arXiv.2203.07474}
\showDOI{\tempurl}
\newblock
\shownote{arXiv:2203.07474 [cs]}.


\bibitem[Iqbal et~al\mbox{.}(2020)]%
        {asu_fpga}
\bibfield{author}{\bibinfo{person}{Odrika Iqbal}, \bibinfo{person}{Saquib Siddiqui}, \bibinfo{person}{Joshua Martin}, \bibinfo{person}{Sameeksha Katoch}, \bibinfo{person}{Andreas Spanias}, \bibinfo{person}{Daniel Bliss}, {and} \bibinfo{person}{Suren Jayasuriya}.} \bibinfo{year}{2020}\natexlab{}.
\newblock \bibinfo{title}{Design and FPGA Implementation of an Adaptive video Subsampling Algorithm for Energy-Efficient Single Object Tracking}.
\newblock , \bibinfo{numpages}{3065-3069}~pages.
\newblock
\urldef\tempurl%
\url{https://doi.org/10.1109/ICIP40778.2020.9191146}
\showDOI{\tempurl}


\bibitem[Khamis et~al\mbox{.}(2018)]%
        {stereonet}
\bibfield{author}{\bibinfo{person}{Sameh Khamis}, \bibinfo{person}{Sean Fanello}, \bibinfo{person}{Christoph Rhemann}, \bibinfo{person}{Adarsh Kowdle}, \bibinfo{person}{Julien Valentin}, {and} \bibinfo{person}{Shahram Izadi}.} \bibinfo{year}{2018}\natexlab{}.
\newblock \showarticletitle{StereoNet: Guided Hierarchical Refinement for Real-Time Edge-Aware Depth Prediction}. In \bibinfo{booktitle}{\emph{Proceedings of the European Conference on Computer Vision (ECCV)}}.
\newblock


\bibitem[Kopf et~al\mbox{.}(2020)]%
        {tiefenrausch}
\bibfield{author}{\bibinfo{person}{Johannes Kopf}, \bibinfo{person}{Kevin Matzen}, \bibinfo{person}{Suhib Alsisan}, \bibinfo{person}{Ocean Quigley}, \bibinfo{person}{Francis Ge}, \bibinfo{person}{Yangming Chong}, \bibinfo{person}{Josh Patterson}, \bibinfo{person}{Jan-Michael Frahm}, \bibinfo{person}{Shu Wu}, \bibinfo{person}{Matthew Yu}, \bibinfo{person}{Peizhao Zhang}, \bibinfo{person}{Zijian He}, \bibinfo{person}{Peter Vajda}, \bibinfo{person}{Ayush Saraf}, {and} \bibinfo{person}{Michael Cohen}.} \bibinfo{year}{2020}\natexlab{}.
\newblock \bibinfo{title}{One Shot 3D Photography}.
\newblock
\newblock
\showeprint[arxiv]{2008.12298}~[cs.CV]


\bibitem[Liu et~al\mbox{.}(2019)]%
        {meta_motiv2}
\bibfield{author}{\bibinfo{person}{Chiao Liu}, \bibinfo{person}{Andrew Berkovich}, \bibinfo{person}{Song Chen}, \bibinfo{person}{Hans Reyserhove}, \bibinfo{person}{Syed~Shakib Sarwar}, {and} \bibinfo{person}{Tsung-Hsun Tsai}.} \bibinfo{year}{2019}\natexlab{}.
\newblock \showarticletitle{Intelligent Vision Systems – Bringing Human-Machine Interface to AR/VR}. In \bibinfo{booktitle}{\emph{2019 IEEE International Electron Devices Meeting (IEDM)}}. \bibinfo{pages}{10.5.1--10.5.4}.
\newblock
\urldef\tempurl%
\url{https://doi.org/10.1109/IEDM19573.2019.8993566}
\showDOI{\tempurl}


\bibitem[Liu et~al\mbox{.}(2022)]%
        {meta_motiv1}
\bibfield{author}{\bibinfo{person}{Chiao Liu}, \bibinfo{person}{Song Chen}, \bibinfo{person}{Tsung-Hsun Tsai}, \bibinfo{person}{Barbara de Salvo}, {and} \bibinfo{person}{Jorge Gomez}.} \bibinfo{year}{2022}\natexlab{}.
\newblock \showarticletitle{Augmented Reality - The Next Frontier of Image Sensors and Compute Systems}. In \bibinfo{booktitle}{\emph{2022 IEEE International Solid-State Circuits Conference (ISSCC)}}, Vol.~\bibinfo{volume}{65}. \bibinfo{pages}{426--428}.
\newblock
\urldef\tempurl%
\url{https://doi.org/10.1109/ISSCC42614.2022.9731584}
\showDOI{\tempurl}


\bibitem[Menze and Geiger(2015)]%
        {kitti2015}
\bibfield{author}{\bibinfo{person}{Moritz Menze} {and} \bibinfo{person}{Andreas Geiger}.} \bibinfo{year}{2015}\natexlab{}.
\newblock \showarticletitle{Object Scene Flow for Autonomous Vehicles}. In \bibinfo{booktitle}{\emph{Conference on Computer Vision and Pattern Recognition (CVPR)}}.
\newblock


\bibitem[{Micron}({[n.\,d.]})]%
        {lpddr5_est}
\bibfield{author}{\bibinfo{person}{{Micron}}.} \bibinfo{year}{[n.\,d.]}\natexlab{}.
\newblock \bibinfo{title}{{LPDDR5} {Memory} {Data} {Sheet}}.
\newblock
\newblock
\urldef\tempurl%
\url{https://www.micron.com/products/dram/lpddr5}
\showURL{%
\tempurl}


\bibitem[{Micron}(2022)]%
        {lpddr4x}
\bibfield{author}{\bibinfo{person}{{Micron}}.} \bibinfo{year}{2022}\natexlab{}.
\newblock \bibinfo{title}{{LPDDR4}/{LPDDR4X} {SDRAM}}.
\newblock
\newblock


\bibitem[Peh and Jerger(2009)]%
        {eecs570_dor}
\bibfield{author}{\bibinfo{person}{Li-Shiuan Peh} {and} \bibinfo{person}{Natalie~Enright Jerger}.} \bibinfo{year}{2009}\natexlab{}.
\newblock \bibinfo{booktitle}{\emph{On-Chip Networks} (\bibinfo{edition}{1st} ed.)}.
\newblock \bibinfo{publisher}{Morgan and Claypool Publishers}.
\newblock
\showISBNx{1598295845}


\bibitem[Pinkham et~al\mbox{.}(2021)]%
        {near_sensor_distributed}
\bibfield{author}{\bibinfo{person}{Reid Pinkham}, \bibinfo{person}{Andrew Berkovich}, {and} \bibinfo{person}{Zhengya Zhang}.} \bibinfo{year}{2021}\natexlab{}.
\newblock \showarticletitle{Near-Sensor Distributed DNN Processing for Augmented and Virtual Reality}.
\newblock \bibinfo{journal}{\emph{IEEE Journal on Emerging and Selected Topics in Circuits and Systems}} \bibinfo{volume}{11}, \bibinfo{number}{4} (\bibinfo{year}{2021}), \bibinfo{pages}{663--676}.
\newblock
\urldef\tempurl%
\url{https://doi.org/10.1109/JETCAS.2021.3121259}
\showDOI{\tempurl}


\bibitem[Pinkham et~al\mbox{.}(2023)]%
        {ansa}
\bibfield{author}{\bibinfo{person}{Reid Pinkham}, \bibinfo{person}{Jack Erhardt}, \bibinfo{person}{Barbara De~Salvo}, \bibinfo{person}{Andrew Berkovich}, {and} \bibinfo{person}{Zhengya Zhang}.} \bibinfo{year}{2023}\natexlab{}.
\newblock \bibinfo{title}{ANSA: Adaptive Near-Sensor Architecture for Dynamic DNN Processing in Compact Form Factors}.
\newblock , \bibinfo{numpages}{1256-1269}~pages.
\newblock
\urldef\tempurl%
\url{https://doi.org/10.1109/TCSI.2022.3228725}
\showDOI{\tempurl}


\bibitem[Redmon and Farhadi(2018)]%
        {redmon_yolov3_2018}
\bibfield{author}{\bibinfo{person}{Joseph Redmon} {and} \bibinfo{person}{Ali Farhadi}.} \bibinfo{year}{2018}\natexlab{}.
\newblock \bibinfo{title}{{YOLOv3}: {An} {Incremental} {Improvement}}.
\newblock
\newblock
\urldef\tempurl%
\url{https://arxiv.org/abs/1804.02767v1}
\showURL{%
\tempurl}


\bibitem[Sandler et~al\mbox{.}(2018)]%
        {mobilenetv2}
\bibfield{author}{\bibinfo{person}{Mark Sandler}, \bibinfo{person}{Andrew Howard}, \bibinfo{person}{Menglong Zhu}, \bibinfo{person}{Andrey Zhmoginov}, {and} \bibinfo{person}{Liang-Chieh Chen}.} \bibinfo{year}{2018}\natexlab{}.
\newblock \showarticletitle{MobileNetV2: Inverted Residuals and Linear Bottlenecks}. In \bibinfo{booktitle}{\emph{Proceedings of the IEEE Conference on Computer Vision and Pattern Recognition (CVPR)}}.
\newblock


\bibitem[Scherer et~al\mbox{.}(2023)]%
        {siracusa}
\bibfield{author}{\bibinfo{person}{Moritz Scherer}, \bibinfo{person}{Manuel Eggimann}, \bibinfo{person}{Alfio~Di Mauro}, \bibinfo{person}{Arpan~Suravi Prasad}, \bibinfo{person}{Francesco Conti}, \bibinfo{person}{Davide Rossi}, \bibinfo{person}{Jorge~Tomás Gómez}, \bibinfo{person}{Ziyun Li}, \bibinfo{person}{Syed~Shakib Sarwar}, \bibinfo{person}{Zhao Wang}, \bibinfo{person}{Barbara~De Salvo}, {and} \bibinfo{person}{Luca Benini}.} \bibinfo{year}{2023}\natexlab{}.
\newblock \showarticletitle{Siracusa: {A} {Low}-{Power} {On}-{Sensor} {RISC}-{V} {SoC} for {Extended} {Reality} {Visual} {Processing} in 16nm {CMOS}}. In \bibinfo{booktitle}{\emph{{ESSCIRC} 2023- {IEEE} 49th {European} {Solid} {State} {Circuits} {Conference} ({ESSCIRC})}}. \bibinfo{pages}{217--220}.
\newblock
\urldef\tempurl%
\url{https://doi.org/10.1109/ESSCIRC59616.2023.10268718}
\showDOI{\tempurl}
\newblock
\shownote{ISSN: 2643-1319}.


\bibitem[Seo et~al\mbox{.}(2022)]%
        {gs_cis1}
\bibfield{author}{\bibinfo{person}{Min-Woong Seo}, \bibinfo{person}{Myunglae Chu}, \bibinfo{person}{Hyun-Yong Jung}, \bibinfo{person}{Suksan Kim}, \bibinfo{person}{Jiyoun Song}, \bibinfo{person}{Daehee Bae}, \bibinfo{person}{Sanggwon Lee}, \bibinfo{person}{Junan Lee}, \bibinfo{person}{Sung-Yong Kim}, \bibinfo{person}{Jongyeon Lee}, \bibinfo{person}{Minkyung Kim}, \bibinfo{person}{Gwi-Deok Lee}, \bibinfo{person}{Heesung Shim}, \bibinfo{person}{Changyong Um}, \bibinfo{person}{Changhwa Kim}, \bibinfo{person}{In-Gyu Baek}, \bibinfo{person}{Doowon Kwon}, \bibinfo{person}{Hongki Kim}, \bibinfo{person}{Hyuksoon Choi}, \bibinfo{person}{Jonghyun Go}, \bibinfo{person}{Jungchak Ahn}, \bibinfo{person}{Jae-Kyu Lee}, \bibinfo{person}{Chang-Rok Moon}, \bibinfo{person}{Kyupil Lee}, {and} \bibinfo{person}{Hyoung-Sub Kim}.} \bibinfo{year}{2022}\natexlab{}.
\newblock \showarticletitle{2.45 e-{RMS} {Low}-{Random}-{Noise}, 598.5 {mW} {Low}-{Power}, and 1.2 kfps {High}-{Speed} 2-{Mp} {Global} {Shutter} {CMOS} {Image} {Sensor} {With} {Pixel}-{Level} {ADC} and {Memory}}.
\newblock \bibinfo{journal}{\emph{IEEE Journal of Solid-State Circuits}} \bibinfo{volume}{57}, \bibinfo{number}{4} (\bibinfo{date}{April} \bibinfo{year}{2022}), \bibinfo{pages}{1125--1137}.
\newblock
\showISSN{1558-173X}
\urldef\tempurl%
\url{https://doi.org/10.1109/JSSC.2022.3142436}
\showDOI{\tempurl}
\newblock
\shownote{Conference Name: IEEE Journal of Solid-State Circuits}.


\bibitem[Shamsafar et~al\mbox{.}(2021)]%
        {mobilestereonet}
\bibfield{author}{\bibinfo{person}{Faranak Shamsafar}, \bibinfo{person}{Samuel Woerz}, \bibinfo{person}{Rafia Rahim}, {and} \bibinfo{person}{Andreas Zell}.} \bibinfo{year}{2021}\natexlab{}.
\newblock \bibinfo{title}{{MobileStereoNet}: {Towards} {Lightweight} {Deep} {Networks} for {Stereo} {Matching}}.
\newblock
\newblock
\urldef\tempurl%
\url{https://arxiv.org/abs/2108.09770v1}
\showURL{%
\tempurl}


\bibitem[Tankovich et~al\mbox{.}(2021)]%
        {hitnet}
\bibfield{author}{\bibinfo{person}{Vladimir Tankovich}, \bibinfo{person}{Christian Hane}, \bibinfo{person}{Yinda Zhang}, \bibinfo{person}{Adarsh Kowdle}, \bibinfo{person}{Sean Fanello}, {and} \bibinfo{person}{Sofien Bouaziz}.} \bibinfo{year}{2021}\natexlab{}.
\newblock \showarticletitle{HITNet: Hierarchical Iterative Tile Refinement Network for Real-time Stereo Matching}. In \bibinfo{booktitle}{\emph{Proceedings of the IEEE/CVF Conference on Computer Vision and Pattern Recognition (CVPR)}}. \bibinfo{pages}{14362--14372}.
\newblock


\bibitem[You et~al\mbox{.}(2022)]%
        {eyecod}
\bibfield{author}{\bibinfo{person}{Haoran You}, \bibinfo{person}{Cheng Wan}, \bibinfo{person}{Yang Zhao}, \bibinfo{person}{Zhongzhi Yu}, \bibinfo{person}{Yonggan Fu}, \bibinfo{person}{Jiayi Yuan}, \bibinfo{person}{Shang Wu}, \bibinfo{person}{Shunyao Zhang}, \bibinfo{person}{Yongan Zhang}, \bibinfo{person}{Chaojian Li}, \bibinfo{person}{Vivek Boominathan}, \bibinfo{person}{Ashok Veeraraghavan}, \bibinfo{person}{Ziyun Li}, {and} \bibinfo{person}{Yingyan Lin}.} \bibinfo{year}{2022}\natexlab{}.
\newblock \bibinfo{title}{EyeCoD: Eye Tracking System Acceleration via Flatcam-Based Algorithm \& Accelerator Co-Design}.
\newblock , \bibinfo{numpages}{13}~pages.
\newblock
\showISBNx{9781450386104}
\urldef\tempurl%
\url{https://doi.org/10.1145/3470496.3527443}
\showDOI{\tempurl}


\bibitem[Zhu et~al\mbox{.}(2022)]%
        {vota}
\bibfield{author}{\bibinfo{person}{Junkang Zhu}, \bibinfo{person}{Wei Tang}, \bibinfo{person}{Ching-En Lee}, \bibinfo{person}{Haolei Ye}, \bibinfo{person}{Eric McCreath}, {and} \bibinfo{person}{Zhengya Zhang}.} \bibinfo{year}{2022}\natexlab{}.
\newblock \showarticletitle{VOTA: A Heterogeneous Multicore Visual Object Tracking Accelerator Using Correlation Filters}.
\newblock \bibinfo{journal}{\emph{IEEE Journal of Solid-State Circuits}} \bibinfo{volume}{57}, \bibinfo{number}{11} (\bibinfo{year}{2022}), \bibinfo{pages}{3490--3502}.
\newblock
\urldef\tempurl%
\url{https://doi.org/10.1109/JSSC.2022.3169946}
\showDOI{\tempurl}


\bibitem[zjjMaiMai(2021)]%
        {tinyhitnet}
\bibfield{author}{\bibinfo{person}{zjjMaiMai}.} \bibinfo{year}{2021}\natexlab{}.
\newblock \bibinfo{title}{TinyHITNet}.
\newblock \bibinfo{howpublished}{\url{https://github.com/zjjMaiMai/TinyHITNet}}.
\newblock


\end{thebibliography}

\end{document}